\def\year{2022}\relax
\documentclass[letterpaper]{article} 
\usepackage{aaai23}  
\usepackage{times}  
\usepackage{helvet}  
\usepackage{courier}  
\usepackage[hyphens]{url}  
\usepackage{graphicx} 
\usepackage{natbib}  
\usepackage{caption} 
\DeclareCaptionStyle{ruled}{labelfont=normalfont,labelsep=colon,strut=off} 
\usepackage{algorithm}
\usepackage{algorithmic}

\usepackage{newfloat}
\usepackage{listings}
\floatstyle{ruled}
\newfloat{listing}{tb}{lst}{}
\floatname{listing}{Listing}

\usepackage[inline]{enumitem}
\usepackage{booktabs}       
\usepackage{dsfont}
\usepackage{amssymb}
\usepackage{amsmath}

\DeclareMathOperator*{\argmax}{arg\,max}
\title{Addressing Distribution Shift at Test Time in Pre-trained Language Models}

\author {
    Ayush Singh, 
    John E. Ortega
}
\affiliations{
    inQbator AI at eviCore Healthcare\\
    Evernorth Health Services\\
    firstname.lastname@evicore.com
}

\begin{document}

\maketitle

\begin{abstract}
State-of-the-art pre-trained language models (PLMs) outperform other models when applied to the majority of language processing tasks. 
However, PLMs have been found to degrade in performance under distribution shift, a phenomenon that occurs when data at test-time does not come from the same distribution as the source training set.
Equally as challenging is the task of obtaining labels in real-time due to issues like long-labeling feedback loops. The lack of adequate methods that address the aforementioned challenges constitutes the need for approaches that continuously adapt the PLM to a distinct distribution. 
Unsupervised domain adaptation adapts a source model to an unseen as well as unlabeled target domain. While some techniques such as data augmentation can adapt models in several scenarios, they have only been sparsely studied for addressing the distribution shift problem. 
In this work, we present an approach (MEMO-CL) that improves the performance of PLMs at test-time under distribution shift. Our approach takes advantage of the latest unsupervised techniques in data augmentation and adaptation to minimize the entropy of the PLM's output distribution. MEMO-CL operates on a batch of augmented samples from a single observation in the test set. 
The technique introduced is unsupervised, domain-agnostic, easy to implement, and requires no additional data. Our experiments result in a 3\% improvement over current test-time adaptation baselines.

\end{abstract}

\noindent In the field of machine learning, the notion that a model will remain accurate over long periods is often accepted as valid. Nonetheless, multiple studies have shown that models do not necessarily perform as well over time when evaluated \citep{quinonero2008dataset,sugiyama2012machine}. 
Lower performance over time can be attributed to several factors; one of the most impacting ones being \emph{distribution shift}.
Distribution shift refers to the change in the underlying semantics of the data that is used to evaluate a model.
In the language domain, the ever-changing nature of language as its spoken or written can be a key contributing factor behind the distribution shift. Consequently, this degradation over time has been found to be prevalent in pre-trained language models (PLMs) as well \citep{lazaridou_mind_2021}. 
The degradation can be harmful to downstream language processing tasks and, thus, novel methods need to be devised to ensure the model performs as intended.

In an attempt to resolve degradation, one could attempt to re-train a PLM \citep{kim_broad_2022}, however, re-training is not only cost prohibitive but could also lead to catastrophic failures \citep{bender2021}. Furthermore, those methods cannot be applied when models are deployed in production i.e. test-time. Even though adapting at test-time does not require re-training, it does require methods to have low run-time complexity. Several works have attempted to address model adaptation to unseen data, albeit without addressing the application of those methods at test-time. 

There are a plethora of augmentation methods that improve performance during the training process \cite{feng_survey_2021}.
Nonetheless, ones that address adaptation during the test-time are less studied, when distribution shift often occurs \cite{wiles2022a}. 
One approach that can be used to increase robustness when deployed to production is called test-time augmentation (TTA \cite{molchanov_greedy_2020}). 
TTA aggregates predictions on a batch of augmented samples generated from a single inference sample to form a final prediction. 
However, TTA approaches only increase robustness in isolation, whereas in the case of distribution shifts, the underlying population itself changes. An ideal technique would not only increase the per-sample robustness of PLM but the entirely shifted distribution as well. 

The primary way of adapting PLMs to a stream of i.i.d. data is known as \textit{continual learning} (CL \cite{cl_dissertation_2022}). 
On the one hand, CL is successful in presence of labels \citep{zhuang_robustly_2021, chawla_quantifying_2021}, but on the other hand it is sub-optimal in the absence of labels, especially under distribution shift. Since label acquisition can be difficult at test-time, some works have used unsupervised methods \citep{ma_domain_2019, wu_domain-agnostic_2021, mishra_surprisingly_2021}, while others used self-supervision  \citep{sohn_fixmatch_2020, sun_test-time_2020, perez-carrasco_con2da_2021, chen_debiased_2022} to adapt a model with unlabeled data. \citep{machireddy_continual_2022, jin_lifelong_2022, cossu_continual_2022} found that continuing to pre-train PLM on the latest data mitigates forgetting as well as adapts the model, however, did not study its efficacy under shift. Moreover, those methods are data-intensive, whereas at test time only a single sample is received. To remedy this, TTA can be used to increase sample efficiency to continually adapt models deployed in production.

In this work, we introduce a technique named MEMO-CL based on the marginal entropy minimization over a single test sample (MEMO) and extend it to the CL paradigm. 
MEMO \cite{zhang_memo_2022} is an approach borrowed from the computer vision literature that increases robustness by doing TTA before adapting the model. However, MEMO only adapts in isolation which limits the benefits from adaptation i.e. it does not address shift nor leverage signal present when more than one observation may suffer from distribution shift. In contrast, our proposed approach continually adapts a model on a stream of test samples. Our alternative method distinctly approaches test-time domain adaptation in a distinct manner from previous methods. MEMO-CL encourages the PLM to predict similarly for semantically similar examples that are augmentations of a single data point. First, it generates synthetic samples from a test sample. Second, it filters these samples using predictions from the base model. Finally, it adapts the model weights before using it for the final prediction. We compare MEMO-CL to an extensive set of baseline methods and find that it improves performance and robustness under distribution shift. 

In order to better illustrate our method, we provide a more detailed description in the following sections. First, we expand on our methodology that adapts models to continually changing data at test-time. Next, we introduce how the quality of the corpus was further improved by selecting informative samples using a margin-based filtering approach. Then, we shed light on the type of data augmentations that MEMO-CL uses. Finally, we share the experiments performed and a discussion on their corresponding outcomes.

\section{Methodology}
\label{method}
In this section, we formally define the MEMO-CL approach. We first describe how our approach is used to adapt PLMs in test-time scenarios using what we call \textit{unsupervised test-time adaptation}. We then cover a method of removing noise produced from the augmentations called \textit{semantic margin-based filtering}. Finally, we show how we use data augmentation to provide samples for MEMO-CL. 

\begin{algorithm}[tb]
\caption{Test-time adaptation via MEMO-CL algorithm}
\label{alg:algorithm}
\textbf{Input}: input $x \in \mathcal{X}$, model $f(x;\theta): \theta \in \Theta$ \\
\textbf{Parameter}: margin $\delta$, number of augmentations $N \in \mathbb{N}$, model weights $\theta \in \Theta$, learning rate $\eta \in \mathbb{R}$, optimizer $\mathcal{O}$ \\
\textbf{Output}: predicted label $\hat{y} \in \mathcal{Y}$
\begin{algorithmic}[1] 
    \STATE Let $i \gets 0, p_{\theta} \gets 0, \tilde{X} \gets \emptyset$
    \WHILE{$i < N$}
        \STATE $a \sim \mathcal{U}(\mathcal{A})$ \hfill \COMMENT$\triangleright$ sample augmentation function (DA)
        \STATE $\tilde{x} \gets a(x)$ \hfill \COMMENT$\triangleright$ augment $x$ (DA)
        \STATE $\tilde{p_{\theta}} \gets f(\tilde{x}|\theta)$ \hfill \COMMENT$\triangleright$ predict using model
        \IF[\hfill $\triangleright$ Eq. \ref{eq:smf} (SMF)]{$\tilde{p_{\theta}}-\delta < \tilde{p_{\theta}} < \tilde{p_{\theta}}+\delta$}
            \STATE $\tilde{X} := \tilde{X} \cup \{\tilde{x}\}$
            \STATE $p_{\theta} := p_{\theta} + \tilde{p_{\theta}}$
            \STATE $i := i + 1$
        \ELSE
            \STATE continue
        \ENDIF
    \ENDWHILE
    \STATE $\ell \gets H(p_{\theta} / N)$ \hfill \COMMENT$\triangleright$ compute loss using Eq. \ref{eq:memo} (UTTA)
    \STATE $\theta' \gets \mathcal{O}(\theta, \eta, \ell) $ \hfill \COMMENT$\triangleright$ update model weights (UTTA)
    \STATE \textbf{return} $\hat{y} \triangleq \argmax_{y \in \mathcal{Y}} f(x|\theta')$
\end{algorithmic}
\end{algorithm}

\subsection{Unsupervised test-time adaptation (UTTA)}
\label{sec:methods:sec:utta}

Algorithm \ref{alg:algorithm} (lines 14--15) illustrates how domain adaptation takes place using MEMO-CL. More formally, given an inference sample $x \in \mathcal{X}$ used to predict label $\hat{y} \in \mathcal{Y}$ using a model $f(x;\theta): \theta \in \Theta$. To do so, from a uniform distribution $\mathcal{U}(\mathcal{A})$ of augmentation functions $a \in \mathcal{A}$, a batch of $N \in \mathbb{N}$ augmented samples $\tilde{x} \leftarrow a_i(x) \, \forall \, i \in [1, N]$ are generated. Therefore, the expectation $\mathds{E}$ of the original model's conditional output distribution $p_{\theta} (y|\tilde{x})$ being consistent across augmentations $\tilde{x}$ is given in Equation \ref{eq:expectation} as follows:
\begin{equation}
    \label{eq:expectation}
    p_{\theta}(y|x) \triangleq \mathds{E}_{\mathcal{U}(\mathcal{A})}[p_{\theta}(y|a(x))] \approx \frac{1}{N} \sum_{n=1}^{N} p_{\theta} (y|\tilde{x}_n)
\end{equation}

One of the main goals of MEMO-CL is to adapt model weights so that the adapted model performs well on unlabeled target data without sacrificing performance.
This is done in a streaming manner which modifies the original work on MEMO \cite{zhang_memo_2022} that limited domain adaptation to per sample basis.
The MEMO-CL approach hence assumes that the distribution shift is not limited to one point in time, in turn covering the entire test-time data set.
MEMO-CL rewards similar predictions from the model that are invariant to perturbations via augmentations by minimizing entropy instead of cross entropy. The idea contrasts the standard form of the cross-entropy loss method which penalizes \textit{confident} yet incorrect predictions. Moreover, minimizing entropy also alleviates the need to get data labeled before performing model adaptation. It does so by minimizing the entropy $H$ of model's marginal output distribution using loss $\ell$ in Equation \ref{eq:memo} as follows: 
\begin{equation}
    \label{eq:memo}
    H(p_{\theta}(y|x)) \triangleq - \sum_{y \in \mathcal{Y}} p_{\theta}(y|x) \log p_{\theta}(y|x)
\end{equation}


\subsection{Semantic margin-based filtering (SMF)}
\label{sec:method:sec:margin}

In Algorithm \ref{alg:algorithm} (lines 6--13), we demonstrate how MEMO-CL handles erroneous augmentations (noisy examples) produced by UTTA. Noisy examples are filtered by keeping only those examples considered crucial for semantic purposes. More formally, semantics preserving augmented sample pool $\tilde{x}$ that are within margin $\delta$ of the prediction probability $p$ of a given sample are kept using an indicator function $\mathds{1}_{SMF}$ defined in Equation \ref{eq:smf}:


\begin{equation}
    \label{eq:smf}
    \mathds{1}_{SMF}(p, \delta) := 
      \begin{cases}
        1, \text{if}\ p-\delta < p < p+\delta \\
        0, \text{otherwise}
      \end{cases} 
\end{equation}

Combining Equation \ref{eq:smf} with Equation \ref{eq:memo} for adaptation, the final loss function can thus be written as in Equation \ref{eq:memo-cl}:
\begin{equation}
    \label{eq:memo-cl}
    \ell(x;\theta) := \mathds{1}_{SMF}(p_{\theta}(y|x), \delta) H(p_{\theta}(y|x))
\end{equation}

\subsection{Data augmentation (DA)}
\label{sec:methods:sec:cntxt_aug}
Another goal of the MEMO-CL approach is to preserve the semantic meaning of the resulting augmented samples. To this end, we follow previous work on
contextual augmentation approaches that provide highly-diverse examples while preserving semantics \cite{feng_survey_2021}. In their work, diversity is defined as the number of semantically-similar modifications performed on the text. Similarly, MEMO-CL uses the following three state-of-the-art context-based augmentation techniques $\mathcal{A}$ at test-time: 
\begin{enumerate}
  \item Synonym replacement using a database of frequent paraphrases \citep{ganitkevitch2013ppdb}.
  \item Synonym replacement by words having similar dense representations using word2vec \citep{mikolov2013distributed}.
  \item Use a standard PLM to paraphrase \citep{kobayashi_contextual_2018, kumar_data_2021, niu-etal-2021-unsupervised}.
\end{enumerate}
A uniform distribution $\mathcal{U}(\mathcal{A})$ from all of the above-listed augmentation techniques $\mathcal{A}$ is created to be used in the MEMO-CL, as defined in Algorithm \ref{alg:algorithm} (lines 3--4).

\section{Experimental settings}
\label{sec:experiments}
In this section, an explanation follows of the experiments performed. The model was adapted with a learning rate $\eta$ of 10e-5 on a machine with 4 NVIDIA V100 GPU, 24 CPU, and 448GB RAM (\textit{Azure} NC24v3). 
Adapting the model takes 10 seconds per sample with $N=20$ augmentations.

\subsubsection{Dataset.} To evaluate our method, we select a dataset that has a significant distributional shift between train and test distribution. The standard for this is WILDS-CivilComments dataset \cite{DBLP:journals/corr/abs-2012-07421} which is a modification of the dataset by \cite{borkan2019nuanced}. This dataset contains 269,038 train and 133,782 test samples along with metadata on belonging to one or more of the 8 sensitive groups. The inputs are sequences with their corresponding binary labels of whether the input is toxic or not.

\subsubsection{Baseline.} The baseline is identical to the one used by \cite{DBLP:journals/corr/abs-2012-07421} which is a DistilBERT \citep{sanh2019distilbert} model fine-tuned on a binary classification task for toxicity detection on the WILDS-CivilComments dataset.

\subsubsection{TTA.} Apart from the baseline, we also compare MEMO-CL with numerous TTA methods. Those are split between \emph{hard} (predicted label) and \emph{soft} (predicted logit) methods. There are numerous ways in which predictions can be aggregated, the primary ones being 
\begin{enumerate*}[label=(\arabic*)]
  \item \emph{majority-voting} uses mode of highest predicted class \citep{wu_domain-agnostic_2021}
  \item \emph{average} simply takes mean of all logits \citep{lu2022improved}
  \item \emph{class weighted} learns optimal weights from a pool of augmentations \citep{shanmugam_better_2021}
\end{enumerate*}. 

\subsubsection{MEMO-CL.} 
To generate contextually augmented samples $\tilde{x}$, the default configuration of \emph{nlpaug} library \citep{ma2019nlpaug} is used to uniformly sample from $a \sim \mathcal{U}(\mathcal{A})$ at test-time. Our method's performance is reported along with improvement achieved by using SMF ($\delta \leftarrow 0.1$).

\subsubsection{Evaluation.}
The original metrics used by \cite{DBLP:journals/corr/abs-2012-07421} included measuring overall \emph{average accuracy} (AA) as well as \emph{worst group accuracy} (WGA) among the 8 different marginalized groups. Additionally, the degree of loss in performance of the approach on the original dataset was measured using \emph{correction to corruption ratio} (CCR). The corruptions are defined as the number of originally correct predictions that were flipped incorrectly, whereas corrections are defined as the number of originally incorrect predictions that were flipped correctly.

\section{Results and conclusion}
\label{results}
Experiments confirm our initial hypothesis of augmenting and then adapting before taking the final prediction. MEMO-CL outperformed baseline as well as TTA methods. Furthermore, by continually learning from incoming data, we notice an additional performance boost over baseline MEMO. The method reduces variance in accuracy not only by minimizing the entropy per sample but that of a stream of augmented samples, thus, increasing label efficiency.
Further reduction in variance was achieved by margin-based filtering (SMF).
The high variance of TTA can be attributed to the augmentations not being perfectly semantic preserving as found by \citep{lu2022improved}. 
The proposed approach also exhibited the highest CCR, however, it was noticed as having higher variance compared to TTA. This adds value by accounting for the inherent noise in the dataset e.g. fixing corruptions by essentially flipping a higher number of corrupt samples.
\begin{table}[h]
  \label{sec:results:table:results}
  \centering
  \small
  \begin{tabular}{llll}
    \toprule
    Model & AA $\uparrow$ & WGA $\uparrow$ & CCR $\uparrow$ \\
    \midrule
    Baseline DistilBERT     & 92.3      & 53.7 &      \\
    TTA majority hard-voting           & - 0.6 (.4)     & - 0.2 (.1)  & 0.91 (.04) \\
    TTA majority soft-voting           & +0.2 (.1)      & +0.4 (.2)  & 1.06 (.05) \\
    TTA average                        & +1.1 (.2)      & +0.8 (.1)  & 1.11 (.03) \\
    TTA class weighted                 & +1.4 (.5)      & +0.6 (.3) & 1.16 (.03) \\
    MEMO                               & +1.6 (.2)      & +0.9 (.1) & 1.12 (.04) \\
    MEMO-CL                               & +2.4 (.3)      & +1.2 (.2) & 1.19 (.09) \\
    MEMO-CL + SMF     & \textbf{+2.9} (.1)      & \textbf{+1.6} (.1) & \textbf{1.21} (.11) \\
    \bottomrule
  \end{tabular}
  \caption{Comparison of MEMO-CL with baseline and existing TTA approaches. The parentheses enclose the standard deviation from 5 runs. See evaluation for details of metrics. }
\end{table}

In this work, a novel technique is presented for addressing the performance degradation of PLMs due to distributional shifts on the fly via augmentation followed by adaptation. 
The technique builds upon recent progress made in unsupervised augmentation, adaptation, and test-time robustness. 
The technique is simple to implement, domain-agnostic and does not require any labeled data.  Experiments show that the MEMO-CL method improves the average and worst-group accuracy over existing approaches.

One potential extension of the present work would be an experimental setup that evaluates the dexterity of the approach in handling not one but multiple simultaneous shifts.
We also posit that learning which type of augmentations will be more suitable for adaptation proposed by \citep{cubuk_autoaugment_2019} is an interesting direction for future work.

\bibliography{aaai23}

\begin{thebibliography}{34}
\providecommand{\natexlab}[1]{#1}

\bibitem[{Bender et~al.(2021)Bender, Gebru, McMillan-Major, and
  Shmitchell}]{bender2021}
Bender, E.~M.; Gebru, T.; McMillan-Major, A.; and Shmitchell, S. 2021.
\newblock On the Dangers of Stochastic Parrots: Can Language Models Be Too Big?
\newblock In \emph{Proceedings of the 2021 ACM Conference on Fairness,
  Accountability, and Transparency}, FAccT '21, 610–623. New York, NY, USA:
  Association for Computing Machinery.
\newblock ISBN 9781450383097.

\bibitem[{Borkan et~al.(2019)Borkan, Dixon, Sorensen, Thain, and
  Vasserman}]{borkan2019nuanced}
Borkan, D.; Dixon, L.; Sorensen, J.; Thain, N.; and Vasserman, L. 2019.
\newblock Nuanced metrics for measuring unintended bias with real data for text
  classification.
\newblock In \emph{Companion Proceedings of The 2019 World Wide Web
  Conference}.

\bibitem[{Chawla, Singh, and Drori(2021)}]{chawla_quantifying_2021}
Chawla, S.; Singh, N.; and Drori, I. 2021.
\newblock Quantifying and Alleviating Distribution Shifts in Foundation Models
  on Review Classification.
\newblock In \emph{NeurIPS 2021 Workshop on Distribution Shifts: Connecting
  Methods and Applications}.

\bibitem[{Chen et~al.(2022)Chen, Jiang, Wang, Wan, Wang, and
  Long}]{chen_debiased_2022}
Chen, B.; Jiang, J.; Wang, X.; Wan, P.; Wang, J.; and Long, M. 2022.
\newblock Debiased {Self}-{Training} for {Semi}-{Supervised} {Learning}.
\newblock ArXiv:2202.07136 [cs].

\bibitem[{Cossu et~al.(2022)Cossu, Tuytelaars, Carta, Passaro, Lomonaco, and
  Bacciu}]{cossu_continual_2022}
Cossu, A.; Tuytelaars, T.; Carta, A.; Passaro, L.~C.; Lomonaco, V.; and Bacciu,
  D. 2022.
\newblock Continual {Pre}-{Training} {Mitigates} {Forgetting} in {Language} and
  {Vision}.
\newblock \emph{ArXiv}.

\bibitem[{Cubuk et~al.(2019)Cubuk, Zoph, Mane, Vasudevan, and
  Le}]{cubuk_autoaugment_2019}
Cubuk, E.~D.; Zoph, B.; Mane, D.; Vasudevan, V.; and Le, Q.~V. 2019.
\newblock {AutoAugment}: {Learning} {Augmentation} {Policies} from {Data}.
\newblock ArXiv:1805.09501 [cs, stat].

\bibitem[{Feng et~al.(2021)Feng, Gangal, Wei, Chandar, Vosoughi, Mitamura, and
  Hovy}]{feng_survey_2021}
Feng, S.~Y.; Gangal, V.; Wei, J.; Chandar, S.; Vosoughi, S.; Mitamura, T.; and
  Hovy, E. 2021.
\newblock A Survey of Data Augmentation Approaches for {NLP}.
\newblock In \emph{Findings of the Association for Computational Linguistics:
  ACL-IJCNLP 2021}, 968--988. Online: Association for Computational
  Linguistics.

\bibitem[{Ganitkevitch, Van~Durme, and
  Callison-Burch(2013)}]{ganitkevitch2013ppdb}
Ganitkevitch, J.; Van~Durme, B.; and Callison-Burch, C. 2013.
\newblock PPDB: The paraphrase database.
\newblock In \emph{Proceedings of the 2013 Conference of the North American
  Chapter of the Association for Computational Linguistics: Human Language
  Technologies}, 758--764.

\bibitem[{Jin et~al.(2022)Jin, Zhang, Zhu, Xiao, Li, Wei, Arnold, and
  Ren}]{jin_lifelong_2022}
Jin, X.; Zhang, D.; Zhu, H.; Xiao, W.; Li, S.-W.; Wei, X.; Arnold, A.; and Ren,
  X. 2022.
\newblock Lifelong {Pretraining}: {Continually} {Adapting} {Language} {Models}
  to {Emerging} {Corpora}.
\newblock In \emph{Proceedings of {BigScience} {Episode} \#5 – {Workshop} on
  {Challenges} \& {Perspectives} in {Creating} {Large} {Language} {Models}},
  1--16. virtual+Dublin: Association for Computational Linguistics.

\bibitem[{Kim et~al.(2022)Kim, Wang, Sclaroff, and Saenko}]{kim_broad_2022}
Kim, D.; Wang, K.; Sclaroff, S.; and Saenko, K. 2022.
\newblock A {Broad} {Study} of {Pre}-training for {Domain} {Generalization} and
  {Adaptation}.
\newblock ArXiv:2203.11819 [cs].

\bibitem[{Kobayashi(2018)}]{kobayashi_contextual_2018}
Kobayashi, S. 2018.
\newblock Contextual {Augmentation}: {Data} {Augmentation} by {Words} with
  {Paradigmatic} {Relations}.
\newblock In \emph{Proceedings of the 2018 {Conference} of the {North}
  {American} {Chapter} of the {Association} for {Computational} {Linguistics}:
  {Human} {Language} {Technologies}, {Volume} 2 ({Short} {Papers})}, 452--457.
  New Orleans, Louisiana: Association for Computational Linguistics.

\bibitem[{Koh et~al.(2020)Koh, Sagawa, Marklund, Xie, Zhang, Balsubramani, Hu,
  Yasunaga, Phillips, Beery, Leskovec, Kundaje, Pierson, Levine, Finn, and
  Liang}]{DBLP:journals/corr/abs-2012-07421}
Koh, P.~W.; Sagawa, S.; Marklund, H.; Xie, S.~M.; Zhang, M.; Balsubramani, A.;
  Hu, W.; Yasunaga, M.; Phillips, R.~L.; Beery, S.; Leskovec, J.; Kundaje, A.;
  Pierson, E.; Levine, S.; Finn, C.; and Liang, P. 2020.
\newblock {WILDS:} {A} Benchmark of in-the-Wild Distribution Shifts.
\newblock \emph{CoRR}, abs/2012.07421.

\bibitem[{Kumar, Choudhary, and Cho(2021)}]{kumar_data_2021}
Kumar, V.; Choudhary, A.; and Cho, E. 2021.
\newblock Data {Augmentation} using {Pre}-trained {Transformer} {Models}.
\newblock ArXiv:2003.02245 [cs].

\bibitem[{Lazaridou et~al.(2021)Lazaridou, Kuncoro, Gribovskaya, Agrawal,
  Liska, Terzi, Gimenez, d'Autume, Kocisky, Ruder, Yogatama, Cao, Young, and
  Blunsom}]{lazaridou_mind_2021}
Lazaridou, A.; Kuncoro, A.; Gribovskaya, E.; Agrawal, D.; Liska, A.; Terzi, T.;
  Gimenez, M.; d'Autume, C. d.~M.; Kocisky, T.; Ruder, S.; Yogatama, D.; Cao,
  K.; Young, S.; and Blunsom, P. 2021.
\newblock Mind the {Gap}: {Assessing} {Temporal} {Generalization} in {Neural}
  {Language} {Models}.
\newblock ArXiv:2102.01951 [cs].

\bibitem[{Lu et~al.(2022)Lu, Shanmugam, Suresh, and Guttag}]{lu2022improved}
Lu, H.; Shanmugam, D.; Suresh, H.; and Guttag, J. 2022.
\newblock Improved Text Classification via Test-Time Augmentation.
\newblock \emph{arXiv preprint arXiv:2206.13607}.

\bibitem[{Ma(2019)}]{ma2019nlpaug}
Ma, E. 2019.
\newblock NLP Augmentation.
\newblock \url{https://github.com/makcedward/nlpaug}.
\newblock Accessed: 2022-10-25.

\bibitem[{Ma et~al.(2019)Ma, Xu, Wang, Nallapati, and Xiang}]{ma_domain_2019}
Ma, X.; Xu, P.; Wang, Z.; Nallapati, R.; and Xiang, B. 2019.
\newblock Domain {Adaptation} with {BERT}-based {Domain} {Classification} and
  {Data} {Selection}.
\newblock In \emph{Proceedings of the 2nd {Workshop} on {Deep} {Learning}
  {Approaches} for {Low}-{Resource} {NLP} ({DeepLo} 2019)}, 76--83. Hong Kong,
  China: Association for Computational Linguistics.

\bibitem[{Machireddy et~al.(2022)Machireddy, Krishnan, Ahuja, and
  Tickoo}]{machireddy_continual_2022}
Machireddy, A.; Krishnan, R.; Ahuja, N.; and Tickoo, O. 2022.
\newblock Continual {Active} {Adaptation} to {Evolving} {Distributional}
  {Shifts}.
\newblock In \emph{2022 {IEEE}/{CVF} {Conference} on {Computer} {Vision} and
  {Pattern} {Recognition} {Workshops} ({CVPRW})}, 3443--3449.
\newblock ISSN: 2160-7516.

\bibitem[{Mikolov et~al.(2013)Mikolov, Sutskever, Chen, Corrado, and
  Dean}]{mikolov2013distributed}
Mikolov, T.; Sutskever, I.; Chen, K.; Corrado, G.~S.; and Dean, J. 2013.
\newblock Distributed representations of words and phrases and their
  compositionality.
\newblock \emph{Advances in neural information processing systems}, 26.

\bibitem[{Molchanov et~al.(2020)Molchanov, Lyzhov, Molchanova, Ashukha, and
  Vetrov}]{molchanov_greedy_2020}
Molchanov, D.; Lyzhov, A.; Molchanova, Y.; Ashukha, A.; and Vetrov, D. 2020.
\newblock Greedy {Policy} {Search}: {A} {Simple} {Baseline} for {Learnable}
  {Test}-{Time} {Augmentation}.
\newblock ArXiv:2002.09103 [cs, stat].

\bibitem[{Niu et~al.(2021)Niu, Yavuz, Zhou, Keskar, Wang, and
  Xiong}]{niu-etal-2021-unsupervised}
Niu, T.; Yavuz, S.; Zhou, Y.; Keskar, N.~S.; Wang, H.; and Xiong, C. 2021.
\newblock Unsupervised Paraphrasing with Pretrained Language Models.
\newblock In \emph{Proceedings of the 2021 Conference on Empirical Methods in
  Natural Language Processing}, 5136--5150. Online and Punta Cana, Dominican
  Republic: Association for Computational Linguistics.

\bibitem[{P{\'e}rez-Carrasco, Protopapas, and
  Cabrera-Vives(2021{\natexlab{a}})}]{mishra_surprisingly_2021}
P{\'e}rez-Carrasco, M.~I.; Protopapas, P.; and Cabrera-Vives, G.
  2021{\natexlab{a}}.
\newblock Con\${\textasciicircum}\{2\}\${DA}: Simplifying Semi-supervised
  Domain Adaptation by Learning Consistent and Contrastive Feature
  Representations.
\newblock In \emph{NeurIPS 2021 Workshop on Distribution Shifts: Connecting
  Methods and Applications}.

\bibitem[{P{\'e}rez-Carrasco, Protopapas, and
  Cabrera-Vives(2021{\natexlab{b}})}]{perez-carrasco_con2da_2021}
P{\'e}rez-Carrasco, M.~I.; Protopapas, P.; and Cabrera-Vives, G.
  2021{\natexlab{b}}.
\newblock Con\${\textasciicircum}\{2\}\${DA}: Simplifying Semi-supervised
  Domain Adaptation by Learning Consistent and Contrastive Feature
  Representations.
\newblock In \emph{NeurIPS 2021 Workshop on Distribution Shifts: Connecting
  Methods and Applications}.

\bibitem[{Pfülb(2022)}]{cl_dissertation_2022}
Pfülb, B. 2022.
\newblock Continual Learning with Deep Learning Methods in an
  Application-Oriented Context.

\bibitem[{Quinonero-Candela et~al.(2008)Quinonero-Candela, Sugiyama,
  Schwaighofer, and Lawrence}]{quinonero2008dataset}
Quinonero-Candela, J.; Sugiyama, M.; Schwaighofer, A.; and Lawrence, N.~D.
  2008.
\newblock \emph{Dataset shift in machine learning}.
\newblock Mit Press.

\bibitem[{Sanh et~al.(2019)Sanh, Debut, Chaumond, and
  Wolf}]{sanh2019distilbert}
Sanh, V.; Debut, L.; Chaumond, J.; and Wolf, T. 2019.
\newblock DistilBERT, a distilled version of BERT: smaller, faster, cheaper and
  lighter.
\newblock \emph{arXiv preprint arXiv:1910.01108}.

\bibitem[{Shanmugam et~al.(2021)Shanmugam, Blalock, Balakrishnan, and
  Guttag}]{shanmugam_better_2021}
Shanmugam, D.; Blalock, D.; Balakrishnan, G.; and Guttag, J. 2021.
\newblock Better {Aggregation} in {Test}-{Time} {Augmentation}.
\newblock ArXiv:2011.11156 [cs].

\bibitem[{Sohn et~al.(2020)Sohn, Berthelot, Li, Zhang, Carlini, Cubuk, Kurakin,
  Zhang, and Raffel}]{sohn_fixmatch_2020}
Sohn, K.; Berthelot, D.; Li, C.-L.; Zhang, Z.; Carlini, N.; Cubuk, E.~D.;
  Kurakin, A.; Zhang, H.; and Raffel, C. 2020.
\newblock {FixMatch}: {Simplifying} {Semi}-{Supervised} {Learning} with
  {Consistency} and {Confidence}.
\newblock ArXiv:2001.07685 [cs, stat].

\bibitem[{Sugiyama and Kawanabe(2012)}]{sugiyama2012machine}
Sugiyama, M.; and Kawanabe, M. 2012.
\newblock \emph{Machine learning in non-stationary environments: Introduction
  to covariate shift adaptation}.
\newblock MIT press.

\bibitem[{Sun et~al.(2020)Sun, Wang, Liu, Miller, Efros, and
  Hardt}]{sun_test-time_2020}
Sun, Y.; Wang, X.; Liu, Z.; Miller, J.; Efros, A.~A.; and Hardt, M. 2020.
\newblock Test-{Time} {Training} with {Self}-{Supervision} for {Generalization}
  under {Distribution} {Shifts}.
\newblock ArXiv:1909.13231 [cs, stat].

\bibitem[{Wiles et~al.(2022)Wiles, Gowal, Stimberg, Rebuffi, Ktena, Dvijotham,
  and Cemgil}]{wiles2022a}
Wiles, O.; Gowal, S.; Stimberg, F.; Rebuffi, S.-A.; Ktena, I.; Dvijotham,
  K.~D.; and Cemgil, A.~T. 2022.
\newblock A Fine-Grained Analysis on Distribution Shift.
\newblock In \emph{International Conference on Learning Representations}.

\bibitem[{Wu, Yue, and Sangiovanni-Vincentelli(2021)}]{wu_domain-agnostic_2021}
Wu, Q.; Yue, X.; and Sangiovanni-Vincentelli, A. 2021.
\newblock Domain-agnostic Test-time Adaptation by Prototypical Training with
  Auxiliary Data.
\newblock In \emph{NeurIPS 2021 Workshop on Distribution Shifts: Connecting
  Methods and Applications}.

\bibitem[{Zhang, Levine, and Finn(2022)}]{zhang_memo_2022}
Zhang, M.; Levine, S.; and Finn, C. 2022.
\newblock {MEMO}: {Test} {Time} {Robustness} via {Adaptation} and
  {Augmentation}.
\newblock ArXiv:2110.09506 [cs].

\bibitem[{Zhuang et~al.(2021)Zhuang, Wayne, Ya, and Jun}]{zhuang_robustly_2021}
Zhuang, L.; Wayne, L.; Ya, S.; and Jun, Z. 2021.
\newblock A {Robustly} {Optimized} {BERT} {Pre}-training {Approach} with
  {Post}-training.
\newblock In \emph{Proceedings of the 20th {Chinese} {National} {Conference} on
  {Computational} {Linguistics}}, 1218--1227. Huhhot, China: Chinese
  Information Processing Society of China.

\end{thebibliography}

\end{document}